\documentclass[a4paper,conference]{IEEEtran}

\usepackage{multirow}
\usepackage{graphicx}
\usepackage{graphics}
\usepackage{epstopdf}

\usepackage{amssymb}% http://ctan.org/pkg/amssymb
\usepackage{pifont}% http://ctan.org/pkg/pifont
\newcommand{\cmark}{\ding{51}}%
\newcommand{\argminE}{\mathop{\mathrm{argmin}}}

\usepackage{amsmath}
\DeclareMathOperator{\sign}{sign}

\newcommand{\ie}{{\em i.e.}}
\newcommand{\eg}{{\em e.g.}}

\newcommand{\eq}[1]{Eq. \ref{#1}}
\newcommand{\Eq}[1]{Eq. (\ref{#1})}

\usepackage{color}
\definecolor{red}{rgb}{1.00,0.20,0.20}
\definecolor{blue}{rgb}{0.20,0.20,1.00}
\definecolor{green}{rgb}{0.00,1.00,0.00}
\begin{document}

\title{Training a Binary Weight Object Detector by Knowledge Transfer for Autonomous Driving}

\author{

\IEEEauthorblockN{Jiaolong Xu\IEEEauthorrefmark{1},
 Peng Wang\IEEEauthorrefmark{2},
  Heng Yang\IEEEauthorrefmark{3} and Antonio M. L\'opez\IEEEauthorrefmark{1}}
  
\IEEEauthorblockA{\IEEEauthorrefmark{1}Computer Vision Center, Autonomous University of Barcelona, Spain\\
{jiaolong, antonio}@cvc.uab.es}
%\IEEEauthorblockA{\IEEEauthorrefmark{2}superwpeng@gmail.com
%\IEEEauthorrefmark{3}chris@ulsee.com}
}


%Telephone: (800) 555--1212, Fax: (888) 555--1212}


\maketitle

\begin{abstract}
Autonomous driving has harsh requirements of small model size and energy efficiency, in order to enable the embedded system to achieve real-time on-board object detection. Recent deep convolutional neural network based object detectors have achieved state-of-the-art accuracy. However, such models are trained with numerous parameters and their high computational costs and large storage prohibit the deployment to memory and computation resource limited systems. Low-precision neural networks are popular techniques for reducing the computation requirements and memory footprint. Among them, binary weight neural network (BWN) is the extreme case which quantizes the float-point into just $1$ bit. BWNs are difficult to train and suffer from accuracy deprecation due to the extreme low-bit representation. To address this problem, we propose a knowledge transfer (KT) method to aid the training of BWN using a full-precision teacher network. We built DarkNet- and MobileNet-based binary weight YOLO-v2 detectors and conduct experiments on KITTI benchmark for car, pedestrian and cyclist detection. The experimental results show that the proposed method maintains high detection accuracy while reducing the model size of DarkNet-YOLO from $257$ MB to $8.8$ MB and MobileNet-YOLO from $193$ MB to $7.9$ MB. 
\end{abstract}


% \fi
\IEEEpeerreviewmaketitle

\section{Introduction}

Autonomous driving requires object detectors to operate on embedded processors to accurately detect cars, pedestrians, cyclists, road signs, and other objects in real-time to ensure safety \cite{wu2016squeezedet}. The state-of-the-art object detectors are trained with deep neural networks (DNNs) which have shown top accuracy for a wide range of computer vision tasks, such like image classification \cite{krizhevsky:2012}, semantic segmentation \cite{long:2015} and object detection \cite{renNIPS15fasterrcnn, Redmon:2016YOLO9000}. The success of deep convolutional neural networks relies on a large amount of labeled training data and powerful computing systems such as GPUs. Deep models generally have high computation cost and require large storage and memory footprints, which prohibits the deployment to resource constrained systems, {\eg} embedded systems. One possible solution is to offload all computations to the cloud but this introduces latency and potentially privacy risks because data is processed remotely. Therefore, developing small size and energy efficient DNN object detector is an emergent task. 

Network compression and prunning have attracted increasing research interests. Currently, there are mainly two categories of techniques to reduce the computational cost of DNNs. One approach is to prune the network by removing redundant weights. A typical way of pruning is to first remove weights with small magnitude and then the network is fine-tuned to recover the lost accuracy \cite{deep_compression:2016}. The second group of techniques is to use low-precision neural networks \cite{INQ:2017, zhou2016dorefa, Ternary:2017NIPS, XNORNET:2016}, since conventional DNNs use float-point format, being power and storage inefficient. Weight quantization has become a popular technique that converts a baseline float-point model into a low-precision representation. The quantization algorithms can be classified into the following groups: fixed-point quantization \cite{courbariaux:2014}, power-of-two quantization \cite{ShiftCNN:2017arXiv}, ternary or binary quantization \cite{Ternary:2017NIPS, XNORNET:2016}. Among them, binary quantization is the extreme case where the float-point weights are represented by only $1$ bit. The binary weight neural networks (BWNs) have favorable largest compression ratio, {\ie} $32\times$, but also sacrifice the most accuracy over the baseline full-precision networks \cite{XNORNET:2016}.

How to improve the accuracies of BWNs has been a challenging problem. Focusing on developing better training strategies,  recently, \cite{Tang:2017} proposed a layer-wise training method and \cite{2017arXiv170904344Z} has done careful analysis on the training tricks, including learning rate, regularization, and activation approximation. Other studies try to compensate the accuracy loss by building more complex model structures to enhance the representation power \cite{DJI:2017, li2017performance}. However, all these methods focus on the BWN itself but neglect its corresponding full-precision counterparts, {\ie} the full-precision counterpart is not involved in the training. In this work, we consider to take the advantage of the high accuracy full-precision model to assist the training of the BWN. 

Our method is inspired by Knowledge Distilling (KD) technique\cite{hinton2015distilling}, which is originally applied to model compression and recently also to the training of low-precision networks \cite{Mishra:2017arXiv}. In this work, we use a full-precision network as teacher network and a BWN as student network. We propose a knowledge transfer method which guides the BWN to mimic the responses of the intermediate layers of the teacher network during the training. As we will show in the experiments, such additional supervision actually improves the convergency of the BWN. As a result, it effectively avoids the accuracy drop in conventional BWN training. Note that the KD technique used in \cite{hinton2015distilling} and \cite{Mishra:2017arXiv} is limited to image classification tasks, since it builds on the last layer of classification network, {\ie} the Softmax layer. Compared to KD, our method is more flexible, since it transfers the knowledge of intermediate layers. As a result, our method can be applied to object detection as well as other tasks. Unlike \cite{DJI:2017, li2017performance}, our method increases neither the model complexity nor the computational cost, and moreover it is easy to implement.

In this work, we present the application of the proposed method to the state-of-the-art YOLO-v2 \cite{Redmon:2016YOLO9000} object detector. However, our method is not limited to this specific detector. It can be applied to any single-stage ({\eg} SSD \cite{SSD:2016}) or two-stage DNN-based detectors ({\eg} Faster R-CNN\cite{renNIPS15fasterrcnn}), or even for other tasks, {\eg} semantic segmentation. 

We use DarkNet and MobileNet \cite{howard2017mobilenets} as the backbones in our binary weight YOLO-v2 detector. The former is the default architecture of YOLO-v2 \cite{Redmon:2016YOLO9000} and the latter is a recent high-accuracy compact network which has much less parameters and is suitable for the deployment on mobile devices. We denote them by DarkNet-YOLO and MobileNet-YOLO respectively. We conduct experiments on KITTI dataset \cite{Geiger:2012}, a defacto benchmark designed for autonomous driving, for car, pedestrian and cyclist detection. Moreover, we also carried experiments on PASCAL VOC dataset \cite{Everingham:2010} which contains $20$ categories of general objects, to verify the generalization of the proposed method . The experimental results show that the proposed method significantly improves the accuracy of BWNs while reduces the model size of DarkNet-YOLO from $257$ MB to $8.8$ MB and MobileNet-YOLO from $193$ MB to $7.9$ MB.
 
\section{Related work}
\textbf{Network quantization and binarization}. Network quantization is an active research topic. Ternary \cite{Ternary:2017NIPS} and binary quantization \cite{XNORNET:2016} aim at quantizing the network at the largest compression ratio. As a consequence, they usually suffer from accuracy degradation. BWN and XNOR-Net \cite{XNORNET:2016} are the most typical binary neural networks. Since XNOR-Net does not only binarize the weight but also the input, its accuracy is much worse than BWN. INQ \cite{INQ:2017} is an incremental network quantization method which progressively quantizes a full-precision network into a low-precision one whose weights are constrained to be either powers of two or zero. A layer-wise network binarization approach is studied in \cite{2017arXiv170904344Z}. In our work, we also use a similar but more efficient stage-wise training strategy consisting of bainarizing groups of layers stage by stage. Compared to \cite{2017arXiv170904344Z}, our training strategy can reduce a lot of training iterations and meanwhile leads to a good accuracy.

\textbf{Knowledge transfer methods}. KT method for model compression could be dated to \cite{model_compression:2006} where a compressed shallow model is trained with pseudo-data labeled by an ensemble of strong classifiers. Recently, \cite{hinton2015distilling} brings it back for DNNs and introduces knowledge distillation (KD). In \cite{Mishra:2017arXiv}, KD is also applied to the training of low-precision networks and several training schemes are studied. Inspired by KD, \cite{Zagoruyko2017AT} proposes to improve the performance of the student CNN by forcing it to mimic the attention maps of the teacher network. In \cite{Zagoruyko2017AT}, it is further studied that KT can be treated as a distribution matching problem, which is similar to domain adaptation \cite{Masana_2017_ICCV, Xu_PAMI:2014}. A new KT loss function is devised to minimize the Maximum Mean Discrepancy (MMD) metric between the feature distributions of student and teacher \cite{Naiyan:2017NIPS}. Our work shares some characteristics of \cite{Zagoruyko2017AT} and \cite{Naiyan:2017NIPS} in the sense that we also transfer knowledge from the intermediate layers. Our KT method is related to \emph{curriculum learning}, as we transfer knowledge from easy tasks first and progressively increase the difficulty in the later stages.

\section{Proposed method}

\begin{figure}
 \begin{minipage}{1.0\linewidth}
  \centering
  \centerline{\includegraphics[width=8cm]{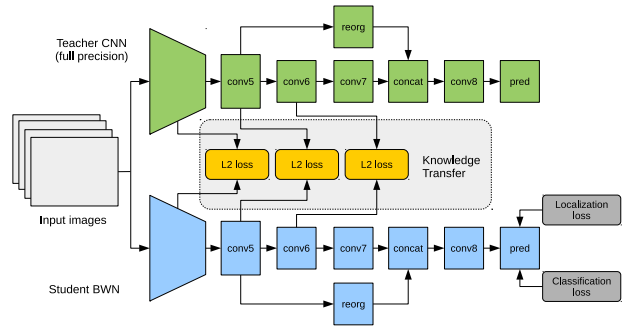}}
  %\vspace{1.5cm}
  %\centerline{(a) With 2 incremental steps}\medskip
\end{minipage}
\caption{The proposed knowledge transfer method for training binary weight YOLO object detector.}
\label{fig:method}
\end{figure}

In this section, we first briefly revisit the binary weight neural networks of \cite{XNORNET:2016}, which is the base of our work. Then we propose several straightforward schemes for efficient training of BWNs, including fine-tuning and state-wise binarization, which are served as strong baselines and can be combined with the proposed KT method. Finally, we elaborate the details of the proposed KT method for training high accuracy BWNs.

\subsection{Binary weight neural network}

BWN is the neural network with binary weights. Let $c$ be the number of channels, $w$ and $h$ the width and the height of the filter respectively, the real valued filter $\textbf{W} \in \mathcal{R}^{c\times w \times h}$ is estimated using a binary filter $\textbf{B} \in \{+1, -1\}^{c\times w \times h}$ and a scaling factor $\alpha \in \mathcal{R}^{+}$, such that $\textbf{W} \approx \alpha \textbf{B}$. The convolution is thus approximated by $\textbf{I} \ast \textbf{W} \approx (\textbf{I} \otimes \textbf{B}) \alpha$, where $\textbf{I}$ is the input data, the symbol $\ast$ represents traditional convolution operation while $\otimes$ indicates the convolution operation only involving additions and subtractions as the weights of the filter are binary. The optimal estimation of $\alpha$ and $\textbf{B}$ is obtained by solving the following optimization:

\begin{equation}
\label{eq:bwn}
\alpha^*, \textbf{B}^* = \argminE_{\alpha, \textbf{B}} \| \textbf{W} - \alpha\textbf{B}\|^2_2
\end{equation}

The solution of \Eq{eq:bwn} is:

\begin{equation}
\label{eq:bwn_sol}
\alpha^* = \frac{1}{n} \|\textbf{W}\|_{l_1}, \textbf{B}^* = \sign(\textbf{W}) ,
\end{equation}

\noindent where $n$ is the number of elements in $\textbf{W}$, $\|\cdot \|_{l_1}$ is the $l_1$ norm and $\sign$ is the sign function which is applied element-wise.

The training of BWN is similar to the ordinary CNNs. In each iteration, given real-valued weights from the previous iteration, the binarized weights are computed accoridng to \eq{eq:bwn_sol}, then the forward propagation of activations and backward propagation of gradients are calculated based on the scaled binary weights. Given the gradient of the scaled binary weights $\tilde{\textbf{W}}$, the gradients of the real-valued weights are calculated by $\frac{\partial C}{\partial \textbf{W}} = \frac{\partial C}{\partial \tilde{\textbf{W}}} (\frac{1}{n} + \frac{\partial \sign}{\partial \textbf{W}}\alpha)$. After that, the real-valued weights are updated by gradient descent. For more details, please refer to \cite{XNORNET:2016}.

\subsection{Fine-tuning and stage-wise binarization}

Before introducing the KT method, we first present several training schemes which can not only improve the training efficiency but also be combined with the KT method. 

Although BWNs and XNOR-Net can be trained from scratch \cite{XNORNET:2016}, fine-tuning with the pre-trained full precision network obtains a faster and better convergence. The effectiveness of such training strategy is also verified in \cite{Mishra:2017arXiv}. In this work, we use the full-precision network to initialize the BWNs and fine-tune from the initialized BWNs.

As observed in \cite{2017arXiv170904344Z}, for BWN, the binarization of the first few layers causes significant accuracy loss while binarizing the last few layers has little effect. A layer-wise priority training strategy is studied in \cite{2017arXiv170904344Z}, where the weights are binarized in reverse order of the layer depth. 
%There are several drawbacks of this method. Firstly, since the accuracy drop of BWN is not exactly proportional to the depth of the network, the training may get stuck at the early stage due to a large loss at certain layer. Secondly, the layer-by-layer binarization requires fine-tuning after each step of binarization, which is time consuming. 
In this work, we propose an analog but more efficient stage-wise training strategy. We first separate the layers into groups and then binarize the groups stage by stage. We also follow a reversed order, {\ie} binarizing from the last group to the first group. Such stage-wise training can be very efficient. According to our experience, it only takes $1$ or $2$ epochs for the first stage training to achieve a comparable accuracy to the full-precision network. The stage-wise binarization can also be interpreted as curriculum learning. For curriculum learning, we first solve easy tasks and then gradually increase the difficulty. Binarizing the whole network from the beginning is much more difficult than stage-wise progressive binarization. The former converges much slower and may get stuck at very bad local minima.

However, fine-tuning and stage-wise binarization make very limited improvement on accuracy as they cannot provide additional supervision. In fact, we found that by carefully tuning the hyper-parameters, it is possible to train BWN from scratch to obtain the same accuracy as these strategies but it requires more iterations. In the next, based on the fine-tuning and stage-wise training, we introduce the proposed KT method which is the key to obtaining high accuracy in this work.

\subsection{Intermediate layer knowledge transfer}

Inspired by the knowledge distillation for model compression \cite{hinton2015distilling} and training low-precision networks \cite{Mishra:2017arXiv}, we propose to transfer knowledge from the intermediate layers of a full-precision network. 

The overall idea is illustrated in \figurename\ref{fig:method}. The pre-trained full-precision teacher network is shown on the top of \figurename\ref{fig:method}, where \emph{reorg} is the feature re-organization layer which stacks the neighborhood features along the channels in order to have the same spatial size as the concatenation layer. On the bottom is the student BWN, whose intermediate layers are connected to the teacher network. The knowledge transfer forces the student to output similar feature responses to the teacher. In this work, we assume the student and the teacher have exact the same network architecture. This allows the student network to be easily initialized by  copying the pre-trained weights of the teacher network. However, our method can be extended to a more general case where the teacher and the student have different architectures but with only some layers in common. In such case, the student can be partially initialized from the corresponding layers of the teacher.

After initialization, we run stage-wise training to binarize the last few layers. As the first stage only takes a few epochs to converge without accuracy loss, we start the iteration of KT training based on the first stage BWN. In the forward propagation, the student network performs binarization according to \Eq{eq:bwn_sol}, and intermediate feature responses of both teacher and the student are calculated. We use the simplest $L_2$ loss to measure the similarity of feature response between the teacher and the student. Given $\textbf{F}_{t_i}$ and $\textbf{F}_{s_i}$ the feature responses of the layer $i$ in the teacher and student network respectively, the $L_2$ loss function minimizes the squared differences between the student (estimated) and teacher (target) features responses: $\mathcal{L}_{2}(\textbf{F}_{t_i}, \textbf{F}_{s_i}) = \|\textbf{F}_{s_i} - \textbf{F}_{t_i}\|_2^2$. Although other complex loss functions can also be employed, {\eg} attention map transfer \cite{Zagoruyko2017AT},
 or MMD \cite{Zagoruyko2017AT}, we find the simple $L_2$ loss works well in practice and leave the investigation of other loss functions as future work. The overall loss of the network can be written as follows:

\begin{equation}
\label{eq:kt_loss}
\mathcal{L}(\textbf{W}) = \lambda_1 \mathcal{L}_{cls}(\textbf{y}_c, \textbf{W}) + \lambda_2 \mathcal{L}_{loc}(\textbf{y}_b, \textbf{W}) + \lambda_3 \sum_{i \in K}\mathcal{L}_{2}(\textbf{F}_{t_i}, \textbf{F}_{s_i}),
\end{equation}

\noindent where $\mathcal{L}_{cls}$ and $\mathcal{L}_{loc}$ are the classification and localization loss respectively for object detection, $\lambda_1$, $\lambda_2$ and $\lambda_3$ are the weights for each loss term, and $K$ is the set of the indices of binary weight convolutional layers. In the backward propagation, the object detection loss, {\ie} localization loss and classification loss, together with the feature matching $L_2$ loss are backward propagated in the student network to compute gradients of the weights. The weights of BWN are then updated using the standard solver {\eg} SGD\cite{bottou2012stochastic} or ADAM\cite{kingma2014adam}. Note that the weights of the teacher network are fixed, thus the computation of backward propagation of the teacher network is not needed. The $L_2$ loss between teacher and student network forces the student to mimic the feature response of the full-precision network. In this way, the pre-trained full-precision network transfers knowledge to the student BWN. The transfered knowledge provides additional supervision to the training of BWN and guides the optimization of BWN along an optimal path. 

\section{Experiments}

We evaluate our proposed method on the KITTI dataset \cite{Geiger:2012}， which is a standard object detection benchmark designed for autonomous driving. The KITTI dataset contains three categories, which are car, pedestrian and cyclist. To evaluate the performance for more categories, we also carry out the experiments on PASCAL VOC dataset \cite{Everingham:2010} which has $20$ categories. The mean average precision (mAP) versus recall criterion is adopted to evaluate the detection performance. 

We use the state-of-the-art object detector YOLO-v2 \cite{Redmon:2016YOLO9000} in our experiments for its efficiency and high accuracy. By default, YOLO uses DarkNet as its backbone network. In addition to that, we also use MobileNet \cite{howard2017mobilenets} as the backbone, which is much more compact than DarkNet 
%($17$ MB versus $80$ MB) 
and meanwhile obtains similar accuracy on image classification. We denote by DarkNet-YOLO and MobileNet-YOLO for these two detectors. The filter sizes of the DarkNet-YOLO and MobileNet-YOLO are listed in Table~\ref{tab:darknet-yolo-arc} and Table~\ref{tab:mobilenet-yolo-arc} respectively, where the bold layer names are the candidate binary layers in our experiments. For each architecture, five types of models are compared, namely  \emph{FP}, \emph{M0}, \emph{M1}, \emph{M2} and \emph{KT}. The definition of the models are as follows. \emph{FP}: the full-precision model; \emph{M0}, \emph{M1} and \emph{M2}: the BWN with the $1$st, $2$nd and $3$rd stage binarization; \emph{KT}: the BWN initialized from \emph{M0} and trained with knowledge transfer. The binary weight layers of each model as well as the model size are shown in Table~\ref{tab:darknet-yolo-arc} and Table~\ref{tab:mobilenet-yolo-arc}.

\begin{table}
 \begin{center}
    \begin{tabular}{  c | c | c | c | c | c }
    Layer & Filter shape                             &\emph{M0}              &\emph{M1}             &\emph{M2}      &\textbf{KT}\\ \hline
    Conv1   & $3\times3\times32$                     &                          &                        &                 &           \\ 
    \textbf{Conv2}   & $3\times3\times32\times64$    &                          &                        &\cmark                 &\cmark\\            
    
    \textbf{Conv3\_1}  & $3\times3\times64\times128$  &                          &                        &\cmark                 &\cmark\\ 
    \textbf{Conv3\_2}  & $1\times1\times128\times64$ &                          &                        &\cmark                 &\cmark\\ 
    \textbf{Conv3\_3} & $3\times3\times64\times128$ &                          &                        &\cmark                 &\cmark\\ 
    
    \textbf{Conv4\_1}  & $3\times3\times128\times256$  &                          &                        &\cmark                 &\cmark\\ 
    \textbf{Conv4\_2}   & $1\times1\times256\times128$ &                          &                        &\cmark                 &\cmark\\ 
    \textbf{Conv4\_3}   & $3\times3\times128\times256$ &                          &                        &\cmark                 &\cmark\\ 
    
    \textbf{Conv5\_1}  & $3\times3\times256\times512$  &                          &\cmark                  &\cmark                 &\cmark\\ 
    \textbf{Conv5\_2}   & $1\times1\times512\times256$ &                          &\cmark                  &\cmark                 &\cmark\\ 
    \textbf{Conv5\_3}   & $3\times3\times256\times512$ &                          &\cmark                  &\cmark                 &\cmark\\ 
    \textbf{Conv5\_4}  & $1\times1\times512\times256$  &                          &\cmark                  &\cmark                 &\cmark\\ 
    \textbf{Conv5\_5}   & $3\times3\times256\times512$ &                          &\cmark                  &\cmark                 &\cmark\\ 
    
    \textbf{Conv6\_1}  & $3\times3\times128\times256$  &                          &\cmark                  &\cmark                 &\cmark\\ 
    \textbf{Conv6\_2}   & $1\times1\times256\times128$ &                          &\cmark                  &\cmark                 &\cmark\\ 
    \textbf{Conv6\_3}   & $3\times3\times128\times256$ &                          &\cmark                  &\cmark                 &\cmark\\ 
    \textbf{Conv6\_4}  & $3\times3\times128\times256$  &                          &\cmark                  &\cmark                 &\cmark\\ 
    \textbf{Conv6\_5}   & $1\times1\times256\times128$ &                          &\cmark                  &\cmark                 &\cmark\\ 
    
    \textbf{Conv7\_1}   & $3\times3\times1024\time1024$     &\cmark               &\cmark             &\cmark                 &\cmark\\ 
    \textbf{Conv7\_2}   & $3\times3\times1024\times1024$    &\cmark               &\cmark             &\cmark                 &\cmark\\ 
    \textbf{Conv8\_1}   & $3\times3\times1024\times1024$    &\cmark               &\cmark             &\cmark                 &\cmark\\ 
    
    Pred   & $1\times1\times1024\times125$         &                          &                        &                 &           \\ \hline
    Size (MB)     &257                       &82                        &12                      &8.8              &8.8 \\
    \end{tabular}
\end{center}
\caption{The binary weight layers and model sizes of \textbf{DarkNet-YOLO} based models.}
\label{tab:darknet-yolo-arc}
\end{table}

\begin{table}
 \begin{center}
    \begin{tabular}{ c | c | c | c | c | c }
    Layer & Filter shape                       &\emph{M0}              &\emph{M1}             &\emph{M2}      &\textbf{KT}\\ \hline

    Conv5\_1\_dw   & $3\times3\times512$              &                  &                    &               &           \\
    \textbf{Conv5\_1\_sep}   & $1\times1\times512\times512$    &                  &                    &\cmark               &\cmark           \\
    Conv5\_2\_dw   & $3\times3\times512$              &                  &                    &               &           \\
    \textbf{Conv5\_2\_sep}   & $1\times1\times512\times512$    &                  &                    &\cmark               &\cmark           \\
    Conv5\_3\_dw   & $3\times3\times512$              &                  &                    &               &           \\
    \textbf{Conv5\_3\_sep}   & $1\times1\times512\times512$    &                  &                    &\cmark               &\cmark           \\
    Conv5\_4\_dw   & $3\times3\times512$              &                  &                    &               &           \\
    \textbf{Conv5\_4\_sep}   & $1\times1\times512\times512$    &                  &                    &\cmark               &\cmark           \\
    Conv5\_5\_dw   & $3\times3\times512$              &                  &                    &               &           \\
    \textbf{Conv5\_5\_sep}   & $1\times1\times512\times512$    &                  &                    &\cmark               &\cmark           \\
    
    Conv5\_6\_dw   & $3\times3\times512$              &                  &                    &               &           \\
    \textbf{Conv5\_6\_sep}   & $1\times1\times512\times1024$    &                  &\cmark                    &\cmark               &\cmark           \\

    Conv6\_dw   & $3\times3\times1024$                 &                  &                    &               &           \\
    \textbf{Conv6\_sep}   & $3\times3\times1024\times1024$      &                  &\cmark                    &\cmark               &\cmark           \\
    
    \textbf{Conv7\_1}   & $3\times3\times1024\time1024$        &\cmark                  &\cmark                    &\cmark               &\cmark           \\
    \textbf{Conv7\_2}   & $3\times3\times1024\times1024$       &\cmark                  &\cmark                    &\cmark               &\cmark           \\
    \textbf{Conv8\_1}   & $3\times3\times1024\times1024$       &\cmark                  &\cmark                    &\cmark               &\cmark           \\
    
    Pred   & $1\times1\times1024\times125$            &                  &                    &               &           \\ \hline
    Size (MB)     &193                       &19                        &13                      &7.9              &7.9 \\
    \end{tabular}
\end{center}
\caption{The binary weight layers and model sizes of \textbf{MobileNet-YOLO} based models.}
\label{tab:mobilenet-yolo-arc}
\end{table}

\subsection{Implementation details}

The proposed methods are implemented using MXNET \cite{chen2015mxnet}. Unless otherwise specified, we use following settings. We take ADAM \cite{kingma2014adam} optimizer with the initial learning rate of $1e-4$. The default batch size is set to $10$ for KITTI and $30$ for PASCAL VOC. The training accuracy is measured by mAP. We train the model for around $500$ and $300$ epochs on KITTI and PASCAL VOC respectively, until the models converge. We use $5$ anchors for the YOLO detectors for all experiments. As it is pointed out in previous literature \cite{XNORNET:2016}, for BWNs, binarizing the first or the last layer will cause significant accuracy drop, thus we keep the first and the last layer in full-precision.

\subsection{KITTI object detection}

\begin{figure}
 \begin{minipage}{1.0\linewidth}
  \centering
  \centerline{\includegraphics[width=8cm]{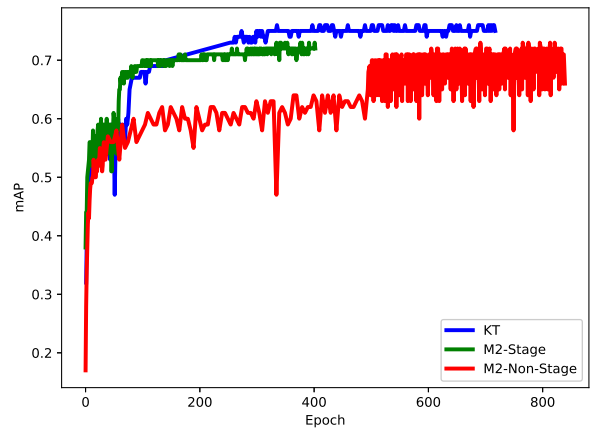}}
  %\vspace{1.5cm}
  %\centerline{(a) With 2 incremental steps}\medskip
\end{minipage}
\caption{Comparison of model convergence rate on KITTI.}
\label{fig:kitti_darknet_val}
\end{figure}

\begin{table}
\begin{center}
\begin{tabular}{c|c|ccc|c}
Method             &mAP     &Pedestrian    &Car            &Cyclist         &Size (MB)\\
\hline
\textbf{FP}        &78      &68            &89             &77                    &257\\
\hline
\textbf{M0}        &78      &67            &89             &76                    &82\\
\textbf{M1}        &76      &65            &88             &75                    &12\\
\textbf{M2}        &72      &58            &87             &72                    &8.8\\
\hline
\textbf{KT}        &76      &64            &88             &76                    &8.8\\
\end{tabular}
\end{center}
\caption{Summary of \textbf{DarkNet-YOLO} detection accuracy and model size on KITTI object detection benchmark.}
\label{tab:kitti_darknet}
\end{table}

\begin{table}
\begin{center}
\begin{tabular}{c|c|ccc|c}
Method             &mAP      &Pedestrian         &Car            &Cyclist   &Size (MB)\\
\hline
\textbf{FP}        &78        &64            &89                 &77            &193\\
\hline
\textbf{M0}        &76        &64            &88                 &76            &19\\
\textbf{M1}        &73        &60            &86                 &72            &13\\
\textbf{M2}        &70        &57            &85                 &67            &7.9\\
\hline
\textbf{KT}        &72        &58            &87                 &72            &7.9\\
\end{tabular}
\end{center}
\caption{Summary of \textbf{MobileNet-YOLO} detection accuracy and model size on KITTI object detection benchmark.}
\label{tab:kitti_mobilenet}
\end{table}

The KITTI dataset contains $7381$ training images. We randomly split it in half as training set and validation set. All images are scaled to canonical size of $1248 \times 384$. We report mean average precision on the validation set.

The results of DarkNet-YOLO and MobileNet-YOLO are shown in Table~\ref{tab:kitti_darknet} and Table~\ref{tab:kitti_mobilenet} respectively. We report mAP across categories as well as the average precision (AP) of each category. The first row of each table shows the results of the full precision model. Both DarkNet and MobileNet based detectors obtain similar accuracy. The second row is the $1$st stage BWN model \emph{M0} which binarizes the $Conv\_7\_1$, $Conv\_7\_2$ and $Conv\_8$ layers. This model has equivalent accuracy as the \emph{FP} but with significant smaller model size, {\ie} $82$ MB versus $257$ MB and $19$ MB versus $193$ MB, which indicates that binarizing the last few layers has little effect to the accuracy. \emph{M1} and \emph{M2} further reduce the model size with more layers binarized. However, there is around $2$ to $5$ percentage points accuracy drop for \emph{M1} and $6$ to $8$ for \emph{M2}. The overall performance of DarkNet-YOLO looks more robust than MobileNet-YOLO which may due to that MobileNet is too compact to be further compressed. The result of the proposed \emph{KT} is shown in the last row. \emph{KT} has the same binarization level as \emph{M2}, {\ie} with the model size of $8.8$ MB and $7.9$ MB for DarkNet-YOLO and MobileNet-YOLO respectively. However, \emph{KT} achieves better accuracy than \emph{M2}, showing the effectiveness of the proposed method.

\figurename~\ref{fig:kitti_darknet_val} depicts the model accuracy on different epochs of the training of DarkNet-YOLO. In this figure, we compare the convergence rate of \emph{KT}, \emph{M2-Stage} and \emph{M2-Non-Stage}. \emph{M2-Stage} is the \emph{M2} BWN fine-tuned from \emph{M1}, {\ie} stage-wise training, while \emph{M2-Non-Stage} is the BWN fine-tuned from \emph{FP}, {\ie} without stage-wise training. \emph{KT} has much faster and better convergence rate than the other two models, which verifies the effectiveness of the additional supervision. \emph{M2-Non-Stage} shows the worst convergence rate. In order to achieve the best accuracy with \emph{M2-Non-Stage}, we have to carefully adjust the learning rate manually. The final accuracy is close to \emph{M2-Stage} but the curve is still unstable, showing the difficulty of convergence. 

\begin{figure*}
 \begin{minipage}{0.5\linewidth}
  \centering
  \centerline{\includegraphics[width=9cm]{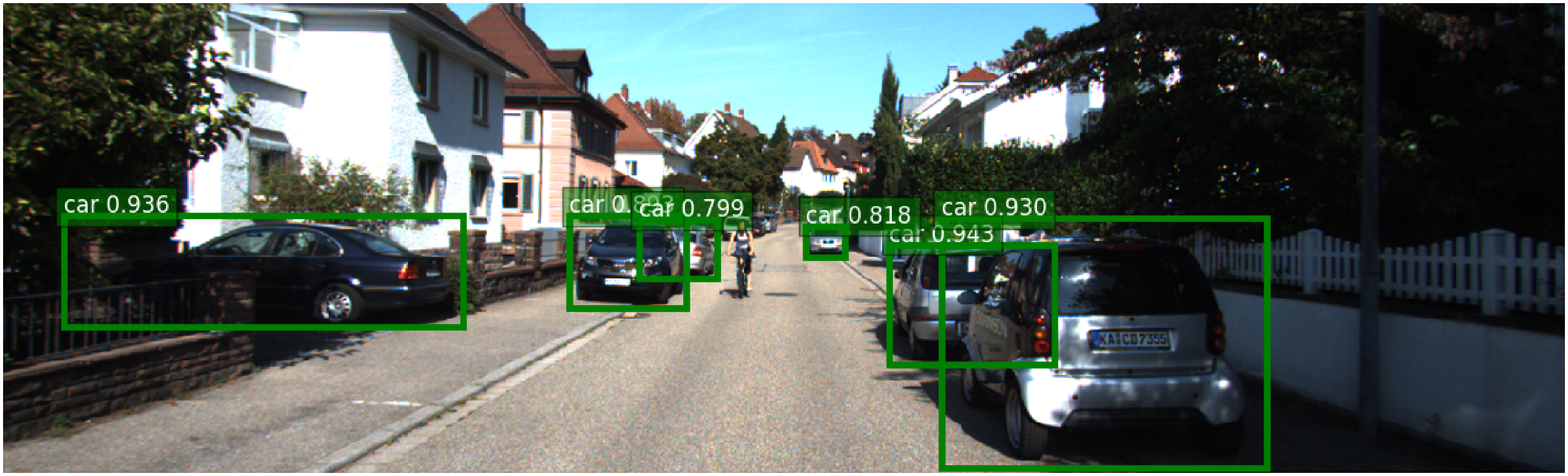}}
  %\vspace{1.5cm}
  %\centerline{(a) With 2 incremental steps}\medskip
\end{minipage}
\begin{minipage}{0.5\linewidth}
  \centering
  \centerline{\includegraphics[width=9cm]{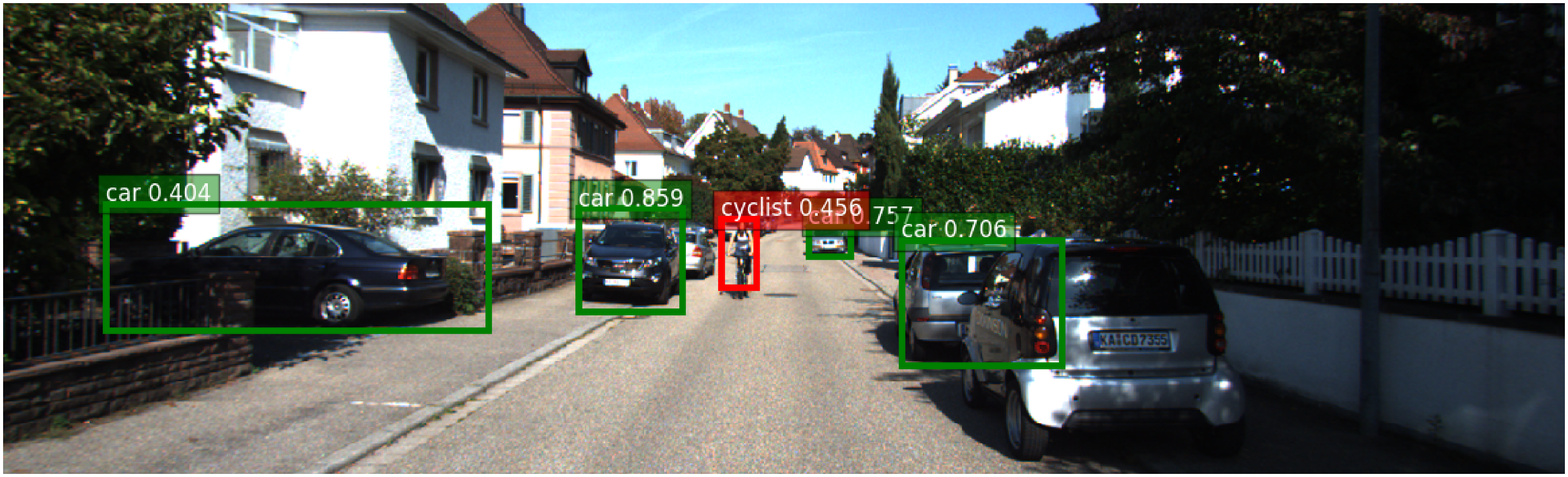}}
  %\vspace{1.5cm}
  %\centerline{(a) With 2 incremental steps}\medskip
\end{minipage}
\begin{minipage}{0.5\linewidth}
  \centering
  \centerline{\includegraphics[width=9cm]{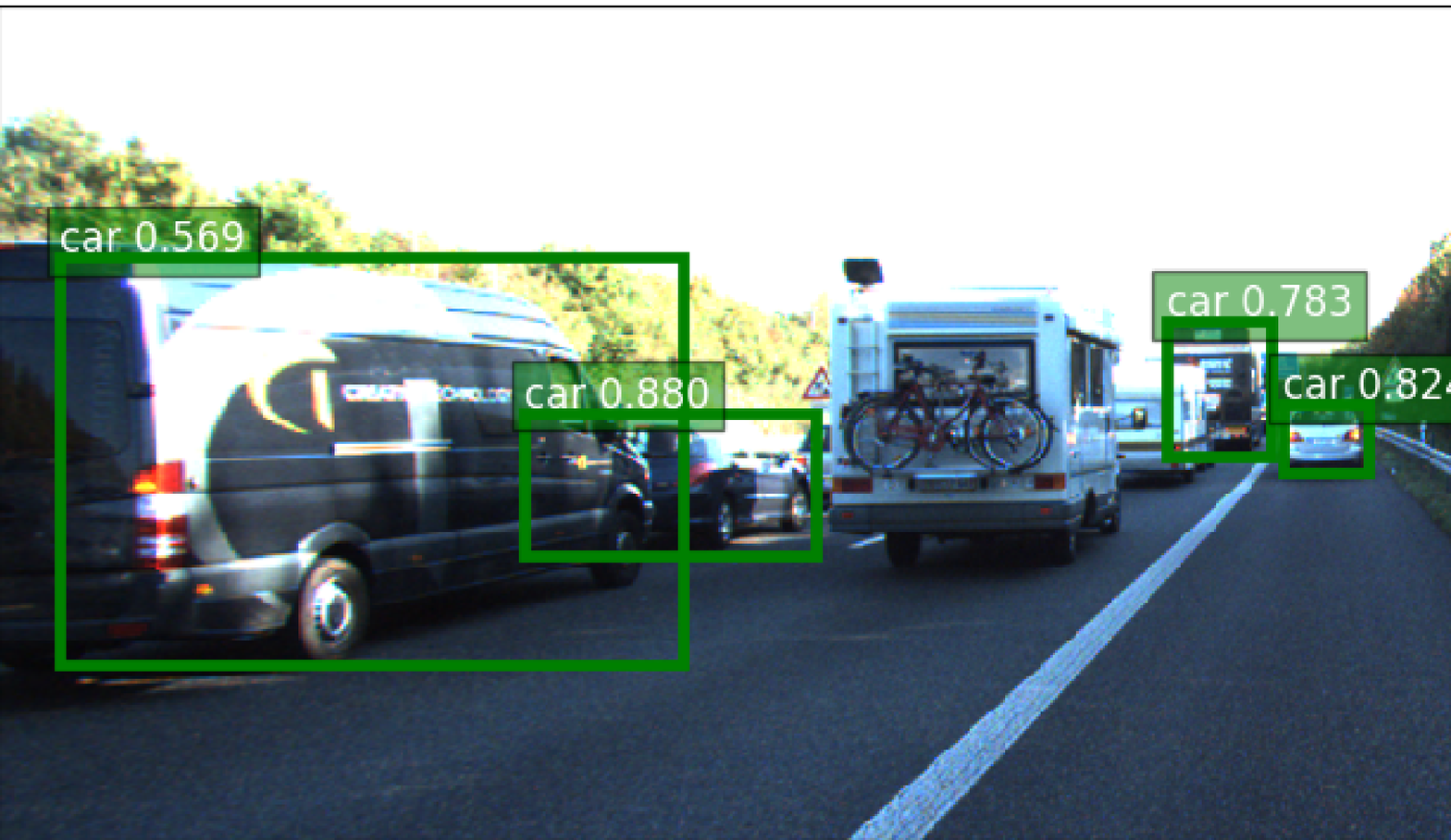}}
  %\vspace{1.5cm}
  %\centerline{(a) With 2 incremental steps}\medskip
\end{minipage}
\begin{minipage}{0.5\linewidth}
  \centering
  \centerline{\includegraphics[width=9cm]{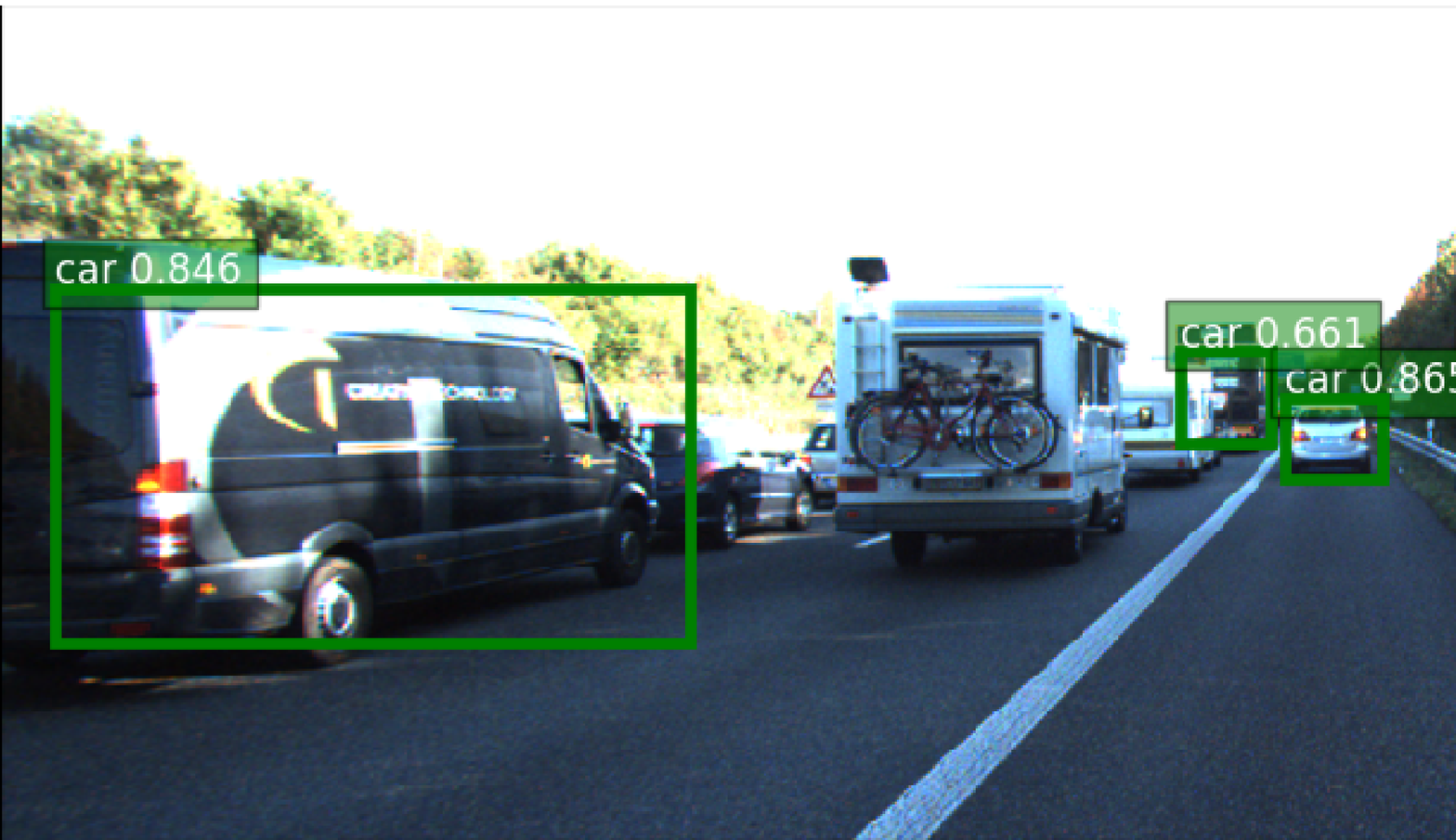}}
  %\vspace{1.5cm}
  %\centerline{(a) With 2 incremental steps}\medskip
\end{minipage}
\begin{minipage}{0.5\linewidth}
  \centering
  \centerline{\includegraphics[width=9cm]{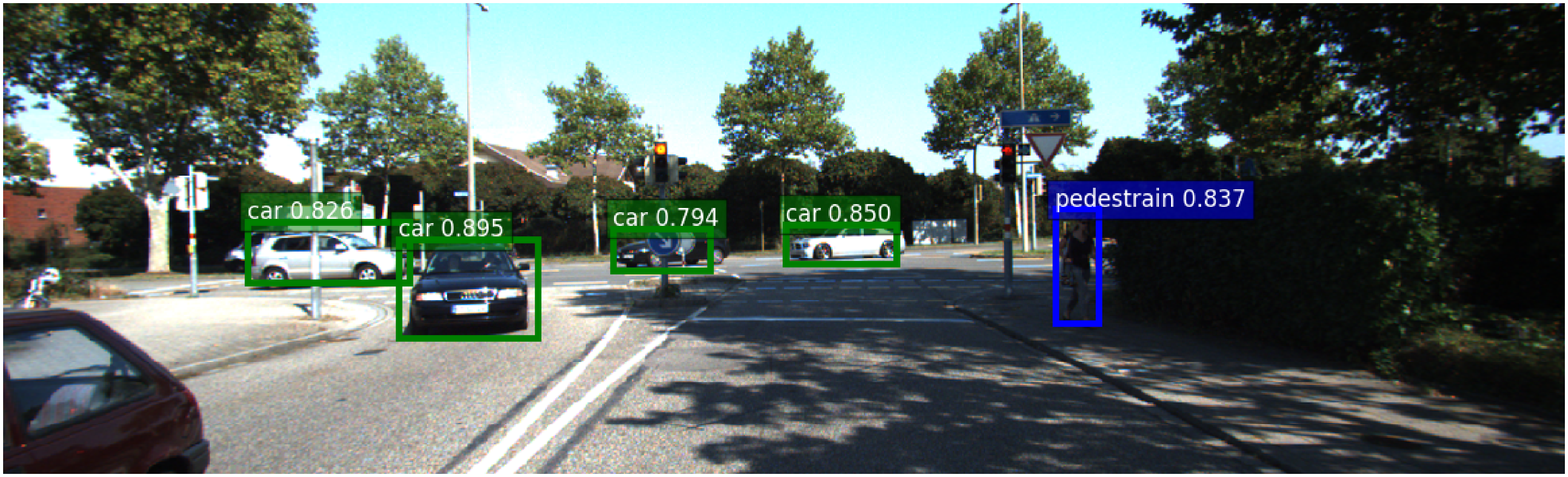}}
  %\vspace{1.5cm}
  %\centerline{(a) With 2 incremental steps}\medskip
\end{minipage}
\begin{minipage}{0.5\linewidth}
  \centering
  \centerline{\includegraphics[width=9cm]{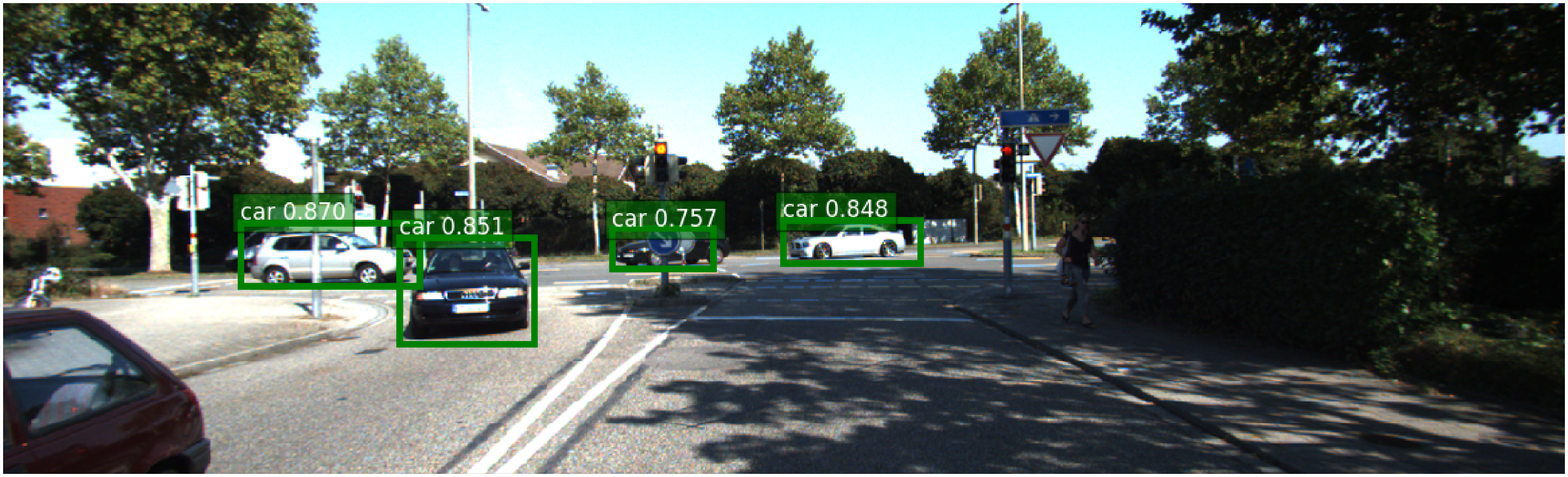}}
  %\vspace{1.5cm}
  %\centerline{(a) With 2 incremental steps}\medskip
\end{minipage}
\caption{Sample detections of \textbf{DarkNet-YOLO} on KITTI dataset (Left: \textbf{FP}, Right: \textbf{KT}).}
\label{fig:demo}
\end{figure*}

\figurename~\ref{fig:demo} shows some typical failure cases of \emph{KT} (right column) when comparing to \emph{FP} (left column). We can see that \emph{FP} achieves better detection accuracy for occluded cars and poorly illumination pedestrians which are typical difficult examples for object detectors. The per-category accuracy in Table~\ref{tab:kitti_darknet} also shows that \emph{KT} losses most of the accuracy on pedestrian detection, but achieves comparable accuracy on car and cyclist detection. The first row of \figurename~\ref{fig:demo} shows that \emph{KT} even occasionally outperforms \emph{FP} for the cyclist detection.

\subsection{PASCAL VOC}

\begin{table*}
\scriptsize
\begin{center}
\begin{tabular}{c|c|p{0.3cm}p{0.3cm}p{0.3cm}p{0.3cm}p{0.3cm}p{0.3cm}p{0.3cm}p{0.3cm}p{0.3cm}p{0.3cm}p{0.3cm}p{0.3cm}p{0.3cm}p{0.3cm}p{0.3cm}p{0.3cm}p{0.3cm}p{0.3cm}p{0.3cm}p{0.3cm}}
Method             &mAP &aero	&bike	&bird	&boat	&bottle	&bus	&car	&cat	&chair	&cow	&table	&dog	&horse	&mbike	&person	&plant	&sheep	&sofa	&train	&tv  \\
\hline
\textbf{FP}        &70 &74 &77 &68 &59 &41 &78 &79 &84 &49 &76 &70 &79 &79 &75 &73 &42 &70 &70 &83 &71        \\
\hline
\textbf{M0}        &69 &71 &78 &68 &57 &40 &77 &79 &82 &48 &73 &69 &78 &79 &76 &72 &41 &66 &66 &84 &71       \\
\textbf{M1}        &66	&69	&75	&65	&57	&42	&75	&76	&78	&46	&67	&63	&71	&77	&77	&72	&37	&69	&61	&81	&71        \\
\textbf{M2}        &62	&64	&73	&56	&48	&28	&73	&74	&74	&39	&64	&65	&66	&77	&70	&67	&34	&57	&62	&77	&62        \\
\hline
\textbf{KT}        &65	&67	&74	&60	&52	&37	&74	&76	&77	&44	&67	&65	&70	&76	&74	&70	&38	&61	&63	&79	&67        \\
\end{tabular}
\end{center}
\caption{Results of \textbf{DarkNet-YOLO} models on PASCAL VOC2007 testing set.}
\label{tab:voc_darknet}
\end{table*}

\begin{table*}
\scriptsize
\begin{center}
\begin{tabular}{c|c|p{0.3cm}p{0.3cm}p{0.3cm}p{0.3cm}p{0.3cm}p{0.3cm}p{0.3cm}p{0.3cm}p{0.3cm}p{0.3cm}p{0.3cm}p{0.3cm}p{0.3cm}p{0.3cm}p{0.3cm}p{0.3cm}p{0.3cm}p{0.3cm}p{0.3cm}p{0.3cm}}
Method             &mAP &aero	&bike	&bird	&boat	&bottle	&bus	&car	&cat	&chair	&cow	&table	&dog	&horse	&mbike	&person	&plant	&sheep	&sofa	&train	&tv  \\
\hline
\textbf{FP}        &69 &71 &78 &70 &56 &41 &75 &76 &82 &42 &73 &69 &77 &79 &74 &71 &43 &65 &62 &81 &70        \\
\hline
\textbf{M0}        &68 &70 &78 &67 &56 &44 &75 &78 &79 &41 &73 &68 &76 &74 &75 &74 &46 &68 &61 &78 &71       \\
\textbf{M1}        &65 &68 &77 &67 &50 &39 &73 &76 &80 &42 &70 &60 &73 &74 &74 &72 &43 &66 &59 &79 &69        \\
\textbf{M2}        &60 &67 &72 &54 &46 &31 &68 &73 &71 &37 &64 &56 &64 &71 &66 &69 &35 &60 &56 &71 &61        \\
\hline
\textbf{KT}        &63 &66 &75 &60 &52 &36 &71 &75 &72 &42 &72 &63 &68 &71 &69 &71 &33 &62 &60 &73 &64        \\
\end{tabular}
\end{center}
\caption{Results of \textbf{MobileNet-YOLO} models on PASCAL VOC2007 testing set.}
\label{tab:voc_mobilenet}
\end{table*}

In this section, we extend the proposed method for general object detection. We conduct the experiments on PASCAL VOC dataset, which contains $20$ categories of common objects. Specifically, we train on VOC2007 trainval and VOC2012 trainval ($16551$ images) and test on VOC2007 test ($4952$ images). For these experiments, all images are re-scaled to canonical size of $416 \times 416$. Table~\ref{tab:voc_darknet} and Table~\ref{tab:voc_mobilenet} present the detection accuracies of DarkNet-YOLO and MobileNet-YOLO models respectively. We obtain similar results as on KITTI dataset. The proposed \emph{KT} method again outperforms \emph{M2} by $3$ percentage points, showing its effectiveness for general object detection.

\section{Conclusion}

In this work, we address the problem of how to train a compact binary weight object detector with high accuracy. First, we reveal that both fine-tuning from full-precision network and the stage-wise binarization are critical and efficient for training BWNs. Moreover, to further improve the model accuracy, we propose to transfer intermediate knowledge from the full-precision network. The experimental results show that the proposed method maintains a high detection accuracy while significantly reduces the model size with a compression rate around $30 \times$.
For the future work, we would like to combine other techniques to further improve the accuracy, {\eg} attention map transfer \cite{Zagoruyko2017AT},
 MMD \cite{Naiyan:2017NIPS} or domain adaptation \cite{Masana_2017_ICCV}.%, and other methods from the domain adaptation area.

\section*{Acknowledgment}

This work is supported by National Natural Science Foundation of China (NSFC, NO. 61601508). Antonio M. L\'opez acknowledges the Spanish project TIN2017-88709-R (Ministerio de Economia, Industria y Competitividad) and the Spanish DGT project SPIP2017-02237, as well as the Generalitat de Catalunya CERCA Program and its ACCIO agency.

{\small
\bibliographystyle{IEEEtran}
\bibliography{references}
}

\end{document}